# A Bayesian Hybrid Parameter-Efficient Fine-Tuning Method for Large Language Models

Yidong Chai[a,b], Yang Liu[a,b], Yonghang Zhou[a,b], Jiaheng Xie[c], Daniel Dajun Zeng[d]

[a] *School of Management, Hefei University of Technology, Hefei, 230009, China*

[b] *Key Laboratory of Process Optimization and Intelligent Decision-making, Ministry of Education, Hefei, 230009, China*

[c] *Department of Accounting & MIS, Lerner College of Business and Economics, University of Delaware, Newark, Delaware, 19716, U.S.*

[d] *Institute of Automation, Chinese Academy of Sciences, Beijing, 100190, China*

## Abstract

**Research Questions and Challenges**: Large Language Models (LLMs) have demonstrated transformative potential in reshaping the world. As these models are pretrained on general corpora, they often require domain-specific fine-tuning to optimize performance in specialized business applications. Due to their massive scale, parameter-efficient fine-tuning (PEFT) methods are widely used to reduce training costs. Among them, hybrid PEFT methods that combine multiple PEFT methods have achieved the best performance. However, existing hybrid PEFT methods face two main challenges when fine-tuning LLMs for specialized applications: 1) relying on point estimates, lacking the ability to quantify uncertainty for reliable decision-making, and 2) struggling to dynamically adapt to newly emerging data, lacking the ability to suit real-world situations. Hence, how to address these challenges has become a pressing need.

**Methodology and Results**: We propose Bayesian Hybrid Parameter-Efficient Fine-Tuning (BH-PEFT), a novel method that integrates Bayesian learning into hybrid PEFT. BH-PEFT combines Adapter, LoRA, and prefix-tuning to fine-tune feedforward layers and attention layers of the Transformer. By modeling the learnable parameters as distributions, BH-PEFT enables uncertainty quantification. We further propose a Bayesian dynamic fine-tuning approach based on BH-PEFT. In each fine-tuning round, the last posterior serves as the prior for the next, allowing for effective adaptation to new data. We evaluated BH-PEFT on various business applications such as sentiment analysis, news categorization, and commonsense reasoning. Experiment results show that our method not only outperforms the existing PEFT baselines, but enables uncertainty quantification for more reliable decisions and improves adaptability in dynamic situations.

**Importance and Implications**: This work contributes to the field of business analytics and data science by proposing a novel BH-PEFT method and a novel dynamic fine-tuning approach. Our method enables uncertainty quantification for more reliable decision-making and effective adaptation in real-world dynamic situations, thus benefiting various business scenarios.

## 1. Introduction

Large Language Models (LLMs) have emerged as a driving force in the rise of generative AI, drawing significant attention from academia, industry, and the media (Chen and Chan 2024, Liu et al. 2025, Ye et al. 2025). Time Magazine described generative AI, including LLMs, as the most significant technological breakthrough since social media, highlighting its potential to reshape various business industries (Chow and Perrigo 2023). For instance, Med-PaLM 2, developed by Google Research and DeepMind, has outperformed medical professionals in answering complex clinical questions and demonstrates strong potential for supporting top-notch diagnostic decision-making (Singhal et al. 2023). Similarly, FinLlama has achieved promising results in financial services, boosting algorithmic trading performance with over 300% cumulative returns and surpassing previous sentiment analysis methods in backtesting (Konstantinidis et al. 2024). Reflecting this momentum, the global LLM market is projected to grow from $6.4 billion in 2024 to $36.1 billion by 2030, with an average annual growth rate of 33.2% (MarketsandMarkets 2024).

However, LLMs are typically pretrained based on generic corpora such as Wikipedia and Common Crawl, and book collections (Zhang et al. 2022, OpenAI 2023). While this provides extensive linguistic knowledge and general reasoning abilities, it does not optimize them for specialized tasks, as generic corpora often lack domain-specific knowledge, task structures, and specialized terminology. This makes them not able to be readily adopted for specialized tasks such as medical diagnosis, legal document analysis, and financial forecasting. As a result, LLM still needs to be fine-tuned to adapt to the specialized tasks. For instance, Med-PaLM 2, as referenced above, a specialized adaptation of PaLM, is fine-tuned on medical datasets to enhance its capability in healthcare-specific tasks and clinical reasoning. Similarly, FinLlama is a specialized adaptation of the LLaMA 2, fine-tuned on labeled financial news datasets to enhance its ability to extract nuanced sentiment signals from financial news and support decision-making in algorithmic trading.

How to fine-tune LLMs has been extensively studied (Mai et al. 2024, Zhu et al. 2025). Full-parameter fine-tuning is initially adopted where all parameters are updated (Devlin et al. 2019). While this can significantly improve LLMs' performance, it is hindered by its high computational costs. For instance, fine-tuning massive models like GPT-3 (175B parameters) demands prohibitive GPU memory and compute resources, limiting practical applications. To mitigate this, parameter-efficient fine-tuning (PEFT) methods have been developed to reduce resource demands while maintaining strong model performance (Han et al. 2024, Albert et al. 2025). Existing PEFT methods include Adapter-based methods, LoRA-based methods, and Prefix-tuning based methods. Adapter-based methods introduce a new module (i.e., adapter) that consists of neural network layers into the Transformer. For instance, the adapters can be inserted sequentially after the multi-head attention layer or after the feed-forward layer of the Transformer (Houlsby et al. 2019). During fine-tuning, only the parameters of the adapters are updated, while the original parameters remain frozen, thereby significantly reducing the number of learnable parameters. LoRA-based methods introduce parallel modules to the original layers (typically the self-attention layers). The output of the introduced parallel module is scaled and added to the original output. During fine-tuning, only the parameters of the introduced parallel modules are updated, while the others remain fixed (Hu et al. 2022). Prefix-tuning based methods prepend learnable vectors (i.e., prefixes) to the model's original input vectors, aiming to update only the prepended vectors while keeping the original input vectors fixed (Li and Liang 2021). By prepending these newly learned vectors to the original input, the LLM is able to perform effectively on a specialized task. While valuable, these methods are limited to focusing on a single aspect of fine-tuning LLMs: Adapter approaches it from the aspect of a sequential module, LoRA from the aspect of a parallel module, and Prefix-tuning from the aspect of prepended vectors. However, these different aspects can be complementary, and as a result, these methods fail to fully leverage the synergies among them.

Hybrid aspect-based methods are proposed to address this limitation. For instance, He et al. (2022) propose to integrate Adapter, LoRA, and Prefix-tuning. They first combine Adapter and LoRA to propose a scaled parallel adapter to fine-tune the feedforward layer, and then use Prefix-tuning to fine-tune the attention layer of the Transformer. While valuable and generally leading to improved performance, the existing hybrid PEFT methods still have two limitations. First, they typically rely on point estimates, which treat model parameters as deterministic point values. Actually, regarding the AI model (including the LLMs) as a model with deterministic point values fails to capture the uncertainty inherent in the model itself (Gal and Ghahramani 2016). Model uncertainty measures a model's confidence in its decision-making. Just like humans, when an AI model responds to a request beyond its knowledge scope, the results are highly uncertain, even if the response happens to be correct (Gawlikowski et al. 2023). Model uncertainty arises when the AI model lacks sufficient knowledge to make a well-informed prediction, yet still produces an output. The inability to capture uncertainty can lead to overconfidence in predictions, increasing the risk of generating plausible yet unreliable or misleading outputs (Huang et al. 2024). Hence, there is a growing need for machine learning models to reflect uncertainty. However, nearly all existing LLMs are trained by treating parameters as deterministic points, thereby failing to account for model uncertainty. Previous studies have demonstrated that capturing uncertainty can enhance decision-making reliability in various applications such as medical diagnosis (Chai et al. 2021), financial risk prediction (Lehrer et al. 2021), and knowledge-intensive question answering (Q. Yang et al. 2023). Therefore, enabling uncertainty quantification in LLMs is crucial for ensuring reliable decision-making on specialized tasks, making it a critical requirement for PEFT methods.

Second, current methods struggle to dynamically adapt to newly emerging data. In practice, data continually accumulates over time, and organizations often need to re-fine-tune previously fine-tuned LLMs to incorporate new data. For instance, in question answering, continual updates are necessary to incorporate new knowledge, as large language models may otherwise recommend outdated answers. When adapting previously fine-tuned LLMs to new data, existing PEFT methods may encounter catastrophic forgetting, leading to the significant loss of previously learned knowledge during re-fine-tuning. Hence, how to effectively fine-tune models in the real-world dynamic situation remains unclear.

To overcome the above two limitations, we propose a novel Bayesian hybrid parameter efficient fine-tuning (BH-PEFT) method. Same as the existing hybrid PEFT methods, our BH-PEFT method also incorporates multiple aspects, such as those of Adapter, LoRA, and Prefix-tuning. We also improve the existing methods by adopting the Bayesian perspective. Specifically, we propose a new Bayesian scaled parallel adapter that combines the aspects of Adapter and LoRA to fine-tune the feedforward layer, and then we propose a Bayesian Prefix-tuning to fine-tune the attention layer of the Transformer. We model the learnable parameters as random variables with a standard Gaussian prior, and infer the posterior distribution given the observed dataset. Since the parameters are distributed, the LLM's outputs consequently become distributed. We therefore quantify the uncertainty by computing the variance of the output distribution. The uncertainty serves as a measure of the reliability of the LLM's outputs, helping to identify plausible but potentially unreliable ones. Furthermore, our Bayesian learning-based method inherently supports dynamic fine-tuning via prior updating, effectively mitigating catastrophic forgetting. Specifically, based on the BH-PEFT, we propose the Bayesian dynamic fine-tuning approach. In each fine-tuning round, the posterior from the last round serves as the prior for the next, enabling the model to adapt efficiently to new data while preserving knowledge from past training. This iterative Bayesian approach makes our method particularly useful for dynamic situations in real world.

We evaluated our method across five benchmark datasets spanning various business applications: sentiment analysis (IMDB and SST2 datasets), news categorization (AG's News dataset), commonsense reasoning (CSQA dataset), and customer satisfaction prediction (Drug Reviews dataset). Our experimental results demonstrate three key findings. First, our method outperformed existing PEFT baselines across all datasets. Second, we observed a negative correlation between uncertainty and correctness (i.e., higher uncertainty indicated lower prediction accuracy), confirming that the quantified uncertainty of our method can effectively flag unreliable LLM outputs to bring about more reliable decision-making. Third, in dynamic fine-tuning, our approach exhibited enhanced stability and superior performance compared to baseline methods.

The key contributions of this study are three-folded. First, we propose a novel BH-PEFT method that integrates Bayesian learning with hybrid PEFT methods. It not only leverages the strengths of different PEFT aspects but also incorporates uncertainty quantification to increase the reliability of the model's decision-making. Second, we develop a new Bayesian dynamic fine-tuning approach that enables LLMs to adapt more effectively to newly emerging data while mitigating the risk of catastrophic forgetting. Third, our method offers a principled way to enhance LLM for various business applications. We validate this via comprehensive experiments on five datasets, and share our code at https://github.com/s22s2s/BH-PEFT to enable broader adoption.

**2. Literature Review on Parameter Efficient Fine-Tuning (PEFT)**

PEFT refers to the methods that adapt LLMs by training only a small subset of parameters, keeping most frozen. This approach drastically cuts computational costs without compromising performance. The major PEFT studies are listed in Table A in Appendix A. PEFT methods can be divided into two types: 1) single-aspect based PEFT, and 2) hybrid-aspect-based PEFT methods.

As the name suggests, single-aspect PEFT methods focus on fine-tuning the Transformer (the architecture is shown in Figure 1(a)) with a single aspect. Different methods focus on different aspects, including the Adapter-based aspect, the LoRA-based aspect, and the Prefix-tuning-based aspect. In the case of Adapter-based fine-tuning, a learnable module (i.e., the adapter) is sequentially introduced into the Transformer, as shown in Figure 1(b). This module is typically inserted after the feedforward layer or the multi-head self-attention layer. For instance, assuming the adapter is a two-layer MLP. Without the adapter, the output is $x$. With the adapter, the output becomes $W_2\sigma(W_1 x)$ where $W_1$ and $W_2$ are the learnable parameters of the two layers of MLP and $\sigma$ is the activation function. During fine-tuning, only the parameters of the adapter (i.e., $W_1$ and $W_2$) are learned, while others remain fixed (Houlsby et al. 2019). LoRA introduces learnable modules into the Transformer in a parallel manner, as shown in Figure 1(c). Assuming the initial weight parameter is $W$, LoRA introduces a weight $W'$ which is equivalent to the product of two low-rank matrices, i.e., $W' = W_{down} W_{up}$ where $W_{down}$ and $W_{up}$ are low-rank matrices. The updated parameter in the fine-tuned LLM becomes $W + \alpha W'$, where $\alpha$ is a scalar hyperparameter that controls the extent of fine-tuning. LoRA is often applied to the query and key matrices in the multi-head self-attention layer. Similarly, only the low-rank matrices $W_{down}$ and $W_{up}$ are learned, while the original model parameters remain fixed during the fine-tuning process (Hu et al. 2022). Prefix-tuning takes a different route by introducing continuous prefix vectors prepended to the original input vectors (typically the keys and values). For instance, assuming the prefix is $P_v$, the new input becomes a vector $[P_v; V]$ where $V$ denotes the original value input. In Prefix-tuning, the prefix vector $P_v$ is usually generated by a prefix encoder (e.g., an MLP), and only the prefix encoder's parameters are trained while other parameters remain fixed (Li and Liang 2021).

**Figure 1. Architecture of Transformer and Different PEFT Aspects**

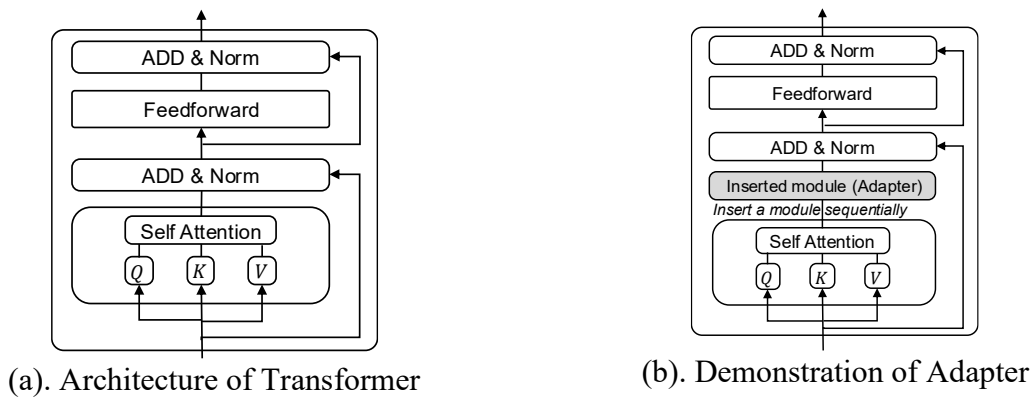

(a). Architecture of Transformer

(b). Demonstration of Adapter

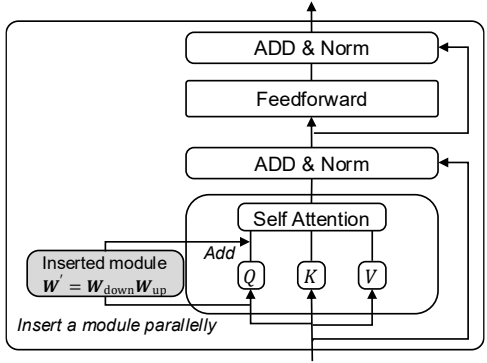 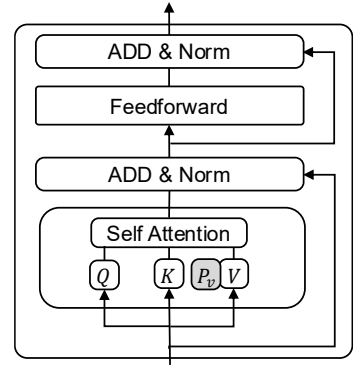

(c). Demonstration of LoRA  (d). Demonstration of Prefix-tuning

Notes: The gray-shaded boxes show the inserted modules. All modules can be placed at alternative positions. For example, the adapter module can also be inserted after the feedforward layer; the LoRA module can be applied to both the key and the feedforward layer; and the prefix module can be introduced for the key. However, for simplicity, we illustrate only one placement. Similar procedures can be applied to other placements.

Building on these basic single-aspect-based PEFT methods, more advanced single-aspect-based PEFT methods have been proposed, as shown in Table A of Appendix A. For instance, COMPACTER extends the adapter by a structured parameterization of adapter weights via Kronecker products (Mahabadi et al. 2021). COMPACTER decomposes the full adapter matrices into sums of Kronecker products between shared slow weights and task-specific fast rank-one matrices. This enables compact and expressive adaptation modules to be inserted into the Transformer. Similarly, Adaptive Prefix Tuning (APT) introduces a dynamic gating mechanism to allocate prefix vectors in a layer-aware and token-sensitive manner. Instead of assigning a fixed-length prefix across all layers, APT adaptively determines the contribution of each pseudo token at different layers by leveraging both fine-grained (token-level) and coarse-grained (layer-level) gates. This allows the model to tailor the prefix capacity based on the characteristics of each layer, leading to more efficient adaptation (Zhang et al. 2023). Albert et al. (2025) proposed a RandLoRA, which extends LoRA by replacing the fully learnable matrices with a combination of non-learnable random matrices and learnable diagonal scaling coefficients. This enables full-rank updates during inference, bringing about improved performance while maintaining efficiency (Albert et al. 2025).

Despite the success of single-aspect-based PEFT methods, they are limited by their focus on isolated single fine-tuning aspect, overlooking synergies and complementarities between different aspects. This limitation has driven the development of hybrid-aspect-based approaches, which integrate multiple aspects to harness their complementary strengths. For instance, Mao et al. (2021a) proposed UniPELT, which integrates adapter, LoRA, and prefix-tuning in the Transformer. Specifically, they apply LoRA to the query and value matrices, use prefix-tuning for keys and values, add an adapter after the feedforward layer along with an additional Add&Norm layer. Meanwhile, they use gating mechanisms to dynamically select different fine-tuning aspects for different tasks (Mao et al. 2021). He et al. (2022) proposed to combine the parallelism and scaling mechanism of LoRA with the adapter, resulting in the Scaled Parallel Adapter (Scaled PA). They then applied the Scaled PA to fine-tune the feedforward layer. Additionally, they used prefix-tuning to fine-tune the key and values of the attention layer. By combining different fine-tuning aspects, hybrid-aspect-based methods typically enhance the performance across a wide range of tasks such as natural language understanding and generation (Mao et al. 2021, He et al. 2022).

While hybrid methods achieve improved performance, they still have two limitations. First, they are limited by point estimates, which treat model parameters as deterministic point values. Denoting the parameters of an LLM as $\boldsymbol{\theta}$, existing methods aim to find an optimal point in the parameter space. However, regarding the machine learning model (including the LLMs) as a model with deterministic point values fails to capture the uncertainty inherent in the model itself. Specifically, given an input example $x$, the output of the model $f_{\boldsymbol{\theta}}(x)$ contains the inherent uncertainty. Moreover, the value of $f_{\boldsymbol{\theta}}(x)$ itself cannot measure such uncertainty. For instance, in classification, $f_{\boldsymbol{\theta}}(x)$ is a predictive probability ranging from 0 to 1 (e.g., from softmax function). It is apparent that when $f_{\boldsymbol{\theta}}(x)$ is far away from 0.5, the $f_{\boldsymbol{\theta}}(x)$ contains less uncertainty. However, previous studies have demonstrated that such quantification is misleading (Gal and Ghahramani 2016). Instead, a more appropriate quantification is Bayesian uncertainty which assumes that $\boldsymbol{\theta}$ is distributed and hence the $f_{\boldsymbol{\theta}}(x)$ is also distributed (Gal and Ghahramani 2016, Maddox et al. 2019, Wang and Yeung 2020). Then, the uncertainty is quantified based on the distribution of $f_{\boldsymbol{\theta}}(x)$ (e.g., based on the variance). Uncertainty is crucial for reliable decision-making, as it helps human experts be aware of how confident the model is in its predictions. Failing to account for uncertainty can have many consequences, such as overconfidence in predictions and a higher risk of generating plausible but misleading outputs (Huang et al. 2024). For instance, a study in financial services found that LLMs used for news-based stock sentiment analysis gave inconsistent results when uncertainty was ignored, leading to unstable investment decisions and higher financial risk (Yu 2023). In contrast, when uncertainty was considered, the model can flag unreliable predictions for human review, resulting in more stable and trustworthy portfolio strategies. However, the existing LLMs are trained by treating model parameters as deterministic point values, and the existing hybrid-aspect-based PEFTs also assume the parameters are deterministic point values. As a result, the obtained LLMs also fail to capture model uncertainty. Hence, it is important to propose a hybrid-aspect-based PEFT method to allow for uncertainty quantification for LLMs.

The second limitation of the existing hybrid PEFT methods is that they struggle to dynamically adapt to new data. In practice, data continually accumulates over time because new data is collected each day of the daily activity. As a result, organizations often need to re-fine-tune previously fine-tuned LLMs to incorporate the information from the new data. However, how to effectively use such new data to fine-tune models is still unclear. Various dynamic fine-tuning methods are proposed, and these methods can be divided into three types: data-pooling-based methods, parameter-initialization-based methods, and data-selection-based methods. The first type of methods (i.e., data-pooling-based) combines the new data and the previous data together to form a data pool, and then uses this data pool to fine-tune the LLMs again (Garg et al. 2024). However, this approach is inefficient given the large amount of previous data and the continued dynamic nature of this process. Moreover, the importance of new data may be undervalued because it is outweighed by the larger amount of previous data. The second type of methods (i.e., parameter-initialization-based) initializes the model with previously trained parameters and then fine-tunes it using the new data. For instance, Soutif et al. (2024) adopt this approach to fine-tune the pre-trained Vision Transformer and ResNet models with the LoRA method. However, the limitation of this approach is that it often leads to catastrophic forgetting, where the model loses previous knowledge while overfitting to the new data, causing performance to decline (Soutif et al. 2024). The third type of studies (i.e., data-selection-based) uses the new data and a subset of previous data to fine-tune. For instance, Song et al. (2023) introduce a dynamic sampling strategy, which hierarchically samples a subset of historical data alongside new data for finetuning (Song et al. 2023). However, deciding which subset to select is usually subjective and heuristic. As a result, the performance is compromised. Hence, the existing methods struggle to dynamically adapt to new data. Hence, how to effectively fine-tune models in a dynamic situation remains unclear.

**2.1 Research Gaps**

In summary, single-aspect PEFT methods have advanced LLM fine-tuning, but hybrid-aspect PEFT methods perform best by combining multiple strengths. Yet, existing hybrid methods face two main limitations. First, they treat model parameters as fixed points, which prevents capturing uncertainty in model outputs and reflecting reliability in decision-making. Second, they struggle to effectively incorporate new data, limiting their practical applicability in real world situations.

To address these two limitations, we propose a novel BH-PEFT method that integrates Bayesian learning into hybrid PEFT, enabling uncertainty quantification. Building on BH-PEFT, we also introduce a Bayesian dynamic fine-tuning approach to account for new data in real world.

## 3. Technical Foundation: Transformer Architecture

Transformer is now the core architecture of LLMs. Transformer models consist of $L$ stacked blocks, each containing a multi-head self-attention layer and a feedforward layer (Vaswani et al. 2017, Zhang et al. 2025). Each of these layers is followed by a residual connection and layer normalization, as illustrated in Figure 1(a). Denoting the input to the self-attention layer as $x_{\text{in}} \in \mathbb{R}^{n \times d}$, where $n$ is the input length and $d$ is the model's hidden dimension. For each head (e.g., $i$-th head), self-attention has three parameter matrices including $W_Q^i \in \mathbb{R}^{d \times d_k}$, $W_K^i \in \mathbb{R}^{d \times d_k}$, and $W_V^i \in \mathbb{R}^{d \times d_v}$ to transform $x_{\text{in}}$ into query, key, and value (denoted as $Q^i, K^i, V^i$). Formally,

$$Q^i = x_{\text{in}} W_Q^i, \quad K^i = x_{\text{in}} W_K^i, \quad V^i = x_{\text{in}} W_V^i \tag{1}$$

Then the self-attention is performed as a weighted sum of value $V^i$ where the weight is determined by the key $K^i$ and the query $Q^i$. Denote the outcome of the $i$-th head as $\text{head}_i$, then,

$$\text{head}_i = \text{Attn}(Q^i, K^i, V^i) = \text{softmax}\left(Q^i {K^i}^{\text{T}} / \sqrt{d_k}\right) V^i \tag{2}$$

where $d_k$ is the dimension of the key. The computation is performed in parallel over $h$ heads. The outputs from all heads are concatenated and projected with a matrix $W_O \in \mathbb{R}^{hd_v \times d}$. Formally,

$$x_{\text{attn}} = \text{Concat}(\text{head}_1, \dots, \text{head}_h) W_O \tag{3}$$

A residual connection is then applied, followed by layer normalization, which normalizes the layer's output to have zero mean and unit variance to stabilize training and improve convergence:

$$x_{\text{rc}} = \text{LayerNorm}(x_{\text{in}} + x_{\text{attn}}) \tag{4}$$

The normalized output $x_{\text{rc}}$ is then passed through a feedforward layer, which typically consists of two linear transformations with a non-linear activation (e.g., ReLU) applied in between:

$$\tilde{x} = \text{ReLU}(x_{\text{rc}} W_1 + b_1) \tag{5}$$

$$x_{\text{ffn}} = \tilde{x} W_2 + b_2 \tag{6}$$

where $W_1$ and $W_2$ are parameter matrices and $b_1, b_2$ are biases of the feedforward layer.

Finally, a residual connection and layer normalization are applied to obtain the output of the current Transformer block:

$$x_{\text{out}} = \text{LayerNorm}(x_{\text{rc}} + x_{\text{ffn}}) \tag{7}$$

$x_{\text{out}}$ is then passed to the next Transformer block, where the same computations are repeated. Each Transformer block has learnable parameters in both the attention layer (e.g., $W_Q^i$ and $W_K^i$) and the feedforward layer (e.g., $W_1$ and $W_2$).

## 4. Our Proposed Bayesian Hybrid Parameter-Efficient Fine-tuning (BH-PEFT) Method

We propose a novel BH-PEFT method that incorporates Bayesian learning into a hybrid PEFT method. Below, we detail the design of BH-PEFT and how the parameters are learned.

### 4.1 The Design of our BH-PEFT Method

Similar to existing hybrid PEFT methods, our BH-PEFT method also incorporates the aspects of Adapter, LoRA, and prefix-tuning. The parameters of the Transformer mainly lie in the attention layer and the feedforward layer. Hence, our BH-PEFT method focuses on these two types of layers.

#### 4.1.1 Fine-tuning for the Multi-Head Self-Attention Layer

We adopt prefix-tuning for the multi-head self-attention layers. Specifically, we introduce a number of $l$ learnable prefix vectors for at each head across all Transformer blocks. Without loss of generality, we demonstrate this for a specific head in a specific Transformer block.

To enhance expressiveness, we follow Li and Liang (2021) where the prefix vectors are not directly optimized but generated by a prefix encoder, featuring a down-projection $W_d^P \in \mathbb{R}^{d \times r_P}$, an up-projection $W_u^P \in \mathbb{R}^{r_P \times d}$ and a non-linear activation function $f$ (e.g., tanh function). $d$ refers to the model's hidden dimension, and $r_P$ is the bottleneck dimension for prefix-tuning. Formally,

$$P_k = W_u^P \cdot f(W_d^P \cdot P_k') \tag{8}$$
$$P_v = W_u^P \cdot f(W_d^P \cdot P_v') \tag{9}$$

where $P_k'$ and $P_v'$ are the inputs to the prefix encoder. The prefix vectors $P_k$ and $P_v$ are split into $h$ head vectors and $P_k^i, P_v^i \in \mathbb{R}^{l \times d/h}$ denote the $i$-th head vector. $P_k^i$ and $P_v^i$ are concatenated with the corresponding key and value vectors to get the outcome of each head:

$$\text{head}_i = \text{Attn}\left(x_{\text{in}} W_Q^i, \text{concat}(P_k^i, x_{\text{in}} W_K^i), \text{concat}(P_v^i, x_{\text{in}} W_V^i)\right) \tag{10}$$

We also introduce Bayesian learning (Guo et al. 2021, Eckman and Henderson 2022, Tan et al. 2024). Specifically, we propose the Bayesian Prefix-tuning which treats $P_k$ and $P_v$ as random variables. Meanwhile, we introduce delta distribution for the original parameters (e.g., $W_K^i$ and $W_V^i$). In this way, we transform the original deterministic LLM into a probabilistic model by representing all parameters as distributions. The $W_K^i$ and $W_V^i$ are fixed, and we aim to learn the distribution of $P_k$ and $P_v$. Since $P_k$ and $P_v$ come from the prefix encoder with parameters $W_u^P$ and $W_d^P$, we treat $W_u^P$ and $W_d^P$ as random variables to make $P_k$ and $P_v$ distributed. Since $W_u^P$ and $W_d^P$ contains continuous values, we assume they follow Gaussian distributions, a common practice in previous studies (Blundell et al. 2015, Wang et al. 2025). Formally:

$$W_d^P \sim \mathcal{N}(\mu_d^P, \Sigma_d^P) \tag{11}$$
$$W_u^P \sim \mathcal{N}(\mu_u^P, \Sigma_u^P) \tag{12}$$

where $\mu_d^P$ and $\mu_u^P$ are the mean, and $\Sigma_d^P$ and $\Sigma_u^P$ are the covariance. Following Blundell et al. (2015), we also adopt the fully factorized Gaussian assumption under which each element is independent:

$$p(W_d^P) = \mathcal{N}\left(W_d^P; \mu_d^P, \text{diag}\left((\sigma_d^P)^2\right)\right) = \prod_{ij} \mathcal{N}\left(W_{d,ij}^P; \mu_{d,ij}^P, (\sigma_{d,ij}^P)^2\right) \tag{13}$$
$$p(W_u^P) = \mathcal{N}\left(W_u^P; \mu_u^P, \text{diag}\left((\sigma_u^P)^2\right)\right) = \prod_{ij} \mathcal{N}\left(W_{u,ij}^P; \mu_{u,ij}^P, (\sigma_{u,ij}^P)^2\right) \tag{14}$$

where $\mu_d^P = [\mu_{d,ij}^P] \in \mathbb{R}^{d \times r_P}$ and $\sigma_d^P = [\sigma_{d,ij}^P] \in \mathbb{R}^{d \times r_P}$, and similarly $\mu_u^P = [\mu_{u,ij}^P] \in \mathbb{R}^{r_P \times d}$ and $\sigma_u^P = [\sigma_{u,ij}^P] \in \mathbb{R}^{r_P \times d}$. These parameters are learned during fine-tuning.

**4.1.2 Fine-tuning for the Feedforward Layer**

He et al. (2022) combined the benefits of LoRA and Adapter and proposes the scaled parallel adapter. It performs well for fine-tuning the feedforward layers. Hence, we also introduce this to fine-tune the feedforward layers in our method. The scaled parallel adapter enhances the original feedforward by an additional non-linear transformation of the input, scaled by a tunable factor $s$:

$$x_{\text{out}} = \text{LayerNorm}(x_{\text{rc}} + x_{\text{ffn}} + s \cdot W_u^A f(W_d^A x_{\text{rc}})) \tag{15}$$

$x_{\text{rc}}$ and $x_{\text{ffn}}$ are the input and output of the original feedforward layer (same notations as thoses in Section 3). $W_d^A \in \mathbb{R}^{d \times r_A}$ and $W_u^A \in \mathbb{R}^{r_A \times d}$ are the down-projection and up-projection matrices. $r_A$ is the bottleneck dimension to fine-tune feedforward layers. $f$ is a non-linear function (e.g., ReLU).

While valuable, the original scaled parallel adapter adopts point estimate and assumes the model parameters are determinstic point values. This limits its capacity to quantify uncertainty to allow for more reliable decision-making. Hence, we introduce Bayesian learning and extend it to a Bayesian scaled parallel adapter. Similar to that in Section 4.1.1, $W_d^A$ and $W_u^A$ are from the fully factorized Gaussian distributions. The corresponding distribution parameters (denote as $\mu_d^A \in \mathbb{R}^{d \times r_A}, \sigma_d^A \in \mathbb{R}^{d \times r_A}, \mu_u^A \in \mathbb{R}^{r_A \times d}$, and $\sigma_u^A \in \mathbb{R}^{r_A \times d}$) are learned during fine-tuning.

**Figure 2. Structure of our Bayesian Hybrid PEFT Method**

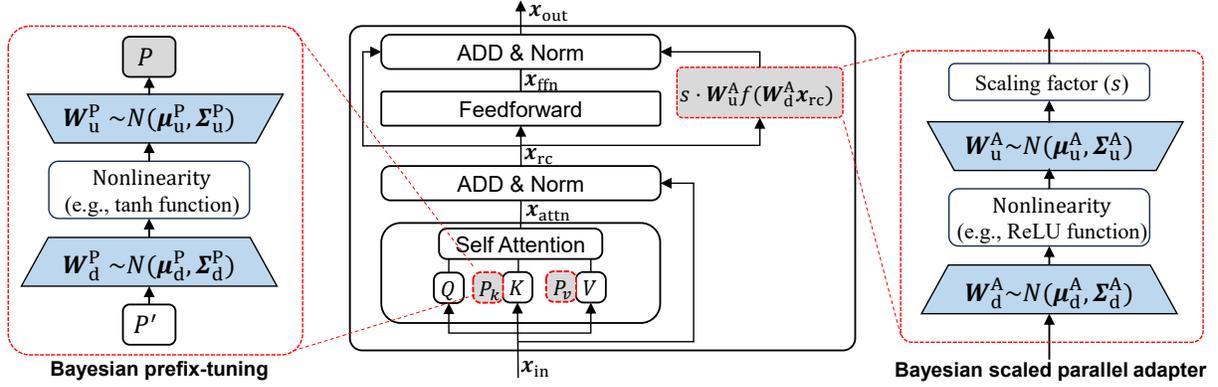

### 4.1.3 Uncertainty Quantification

Now, assuming we have gotten the distributions over the learnable parameters (details of how to learn them will be shown later) including $W_d^A$, $W_u^A$, $W_d^P$ and $W_u^P$, we can then use them to obtain the model's prediction and the associated uncertainty. Formally, for a given input $x^{new}$, the model's prediction is obtained by integrating over the distributions of the parameters as:

$$p(y|x^{new}) = \int p(y|x^{new}, W_d^A, W_u^A, W_d^P, W_u^P) p(W_d^A) p(W_u^A) p(W_d^P) p(W_u^P) dW_d^A dW_u^A dW_d^P dW_u^P \quad (16)$$

This is actually the expected value of the prediction:

$$p(y|x^{new}) = \mathbb{E}_{W_d^A, W_u^A, W_d^P, W_u^P} p(y|x^{new}, W_d^A, W_u^A, W_d^P, W_u^P) \quad (17)$$

For the uncertainty, we quantify it with the variance of the output distribution, a common practice in Bayesian deep learning (Gal and Ghahramani 2016). Specifically, the uncertainty is:

$$u(y|x^{new}) = \text{Var}_{W_d^A, W_u^A, W_d^P, W_u^P} p(y|x^{new}, W_d^A, W_u^A, W_d^P, W_u^P) \quad (18)$$

By modeling the parameters as distributions, we not only obtain the prediction but also the uncertainty behind such prediction. Though the initial parameters of the LLMs are deterministic points and fail to provide uncertainty, our BH-PEFT method can alleviate such problem, allowing us to gauge the reliability of the LLM's output to help us get an informed and reliable decision.

### 4.2 The Learning of our BH-PEFT Method

#### 4.2.1 The Objective of the Learning

LLMs typically comprise multiple Transformer blocks and our BH-PEFT method can be applied across different blocks. However, for simplicity, we show it on a single block.

The objective of learning is to determine the distributions of parameters including $W_d^A$, $W_u^A$, $W_d^P$ and $W_u^P$. As a common practice (Jordan et al. 1999, Blei et al. 2017), we set their prior and infer their posteriors based on the dataset $\mathcal{D} = \{(x_1, y_1), (x_2, y_2), \ldots, (x_N, y_N)\}$. The posterior is obtained by $p(W_d^A, W_u^A, W_d^P, W_u^P | \mathcal{D}) = p(W_d^A, W_u^A, W_d^P, W_u^P) p(\mathcal{D} | W_d^A, W_u^A, W_d^P, W_u^P) / p(\mathcal{D})$. Since $p(\mathcal{D}) = \int p(\mathcal{D} | W_d^A, W_u^A, W_d^P, W_u^P) dW_d^A dW_u^A dW_d^P dW_u^P$, it is intractable, making the posterior hard to compute. To address this, we adopt the variational inference approach which approximates the posterior with a tractable variational distribution $q(W_d^A, W_u^A, W_d^P, W_u^P)$. To reduce the discrepancy due to the approximation, we hope to reduce the distance between variational distribution and the posterior distribution (Liu et al. 2021, Yin et al. 2021). We measure the distance with Kullback–Leibler (KL) divergence (Cheng et al. 2021, Wei et al. 2022). Formally,

$$q^*(W_d^A, W_u^A, W_d^P, W_u^P) = \operatorname{argmin}_q \operatorname{KL}[q(W_d^A, W_u^A, W_d^P, W_u^P) \| p(W_d^A, W_u^A, W_d^P, W_u^P | \mathcal{D})] \quad (19)$$

The KL divergence can be further derived as (details in Appendix B):

$$\operatorname{KL} = \mathbb{E}_q[\log q(W_d^A, W_u^A, W_d^P, W_u^P) - \log p(\mathcal{D}, W_d^A, W_u^A, W_d^P, W_u^P)] + \log p(\mathcal{D}) \quad (20)$$

Since the KL divergence is non-negative, we derive a lower bound on the log-likelihood of the dataset $\log p(\mathcal{D})$ (called evidence lower bound, ELBO), as follows:

$$\operatorname{ELBO} \triangleq \log p(\mathcal{D}) - \operatorname{KL} = \mathbb{E}_q[\log p(\mathcal{D}, W_d^A, W_u^A, W_d^P, W_u^P) - \log q(W_d^A, W_u^A, W_d^P, W_u^P)] \quad (21)$$

Since $p(\mathcal{D}, W_d^A, W_u^A, W_d^P, W_u^P) = p(\mathcal{D} | W_d^A, W_u^A, W_d^P, W_u^P) p(W_d^A, W_u^A, W_d^P, W_u^P)$, the ELBO can be further decomposed into (details in Appendix B):

$$\operatorname{ELBO} = \mathbb{E}_q[\log p(\mathcal{D} | W_d^A, W_u^A, W_d^P, W_u^P)] - \operatorname{KL}(q(W_d^A, W_u^A, W_d^P, W_u^P) \| p(W_d^A, W_u^A, W_d^P, W_u^P)) \quad (22)$$

Same as previous studies (Kingma and Welling 2013), we assume the prior for each parameter as a standard Gaussian. Hence, we have $p(W_d^A, W_u^A, W_d^P, W_u^P) = p(W_d^A) p(W_u^A) p(W_d^P) p(W_u^P)$. Meanwhile, we follow mean-field assumption (Blei et al. 2017) and assume the variational distribution is also factorized as $q(W_d^A, W_u^A, W_d^P, W_u^P) = q(W_d^A) q(W_u^A) q(W_d^P) q(W_u^P)$. Given the independence of the parameters under the mean-field assumption and the factorization of the prior and variational distributions, the ELBO becomes:

$$\begin{aligned} \operatorname{ELBO} = &\, \mathbb{E}_q[\log p(\mathcal{D} | W_d^A, W_u^A, W_d^P, W_u^P)] \\ &- \big( \operatorname{KL}(q(W_d^A) \| p(W_d^A)) + \operatorname{KL}(q(W_u^A) \| p(W_u^A)) + \operatorname{KL}(q(W_d^P) \| p(W_d^P)) \\ &+ \operatorname{KL}(q(W_u^P) \| p(W_u^P)) \big) \end{aligned} \quad (23)$$

For each parameter matrix ($W_d^A, W_u^A, W_d^P$, or $W_u^P$), the variational distribution is a Gaussian distribution. Formally, $q(W_d^A) = \mathcal{N}(\mu_d^A, \Sigma_d^A)$, $q(W_u^A) = \mathcal{N}(\mu_u^A, \Sigma_u^A)$, $q(W_d^P) = \mathcal{N}(\mu_d^P, \Sigma_d^P)$, and $q(W_u^P) = \mathcal{N}(\mu_u^P, \Sigma_u^P)$. We hope to learn the parameters of these Gaussian distributions, i.e., $\mu_d^A, \Sigma_d^A, \mu_u^A, \Sigma_u^A, \mu_d^P, \Sigma_d^P, \mu_u^P, \Sigma_u^P$. We learn these parameters by maximizing ELBO, i.e., the objective in the learning task. Next, we will analyze the two terms of ELBO in Equation (23).

### 4.2.2 The First Term of the Objective

Computing the first term of the ELBO involves integrating over the variational distribution, which is typically computationally intractable. To address this, we employ Monte Carlo method. Meanwhile, we also use the reparameterization trick to enable backpropagation of gradients (Kingma and Welling 2013). Without loss of generality, we show this for $W_d^A$ as follows:

$$W = \mu + \sigma \odot \epsilon, \epsilon \sim \mathcal{N}(0, I) \tag{24}$$

$\mu$ and $\sigma$ denotes the mean and standard deviation of the variational distribution. $\odot$ is element-wise multiplication. $\epsilon$ is a random noise matrix drawn from the standard Gaussian. In this way, we get a sample of $W_d^A$. By sampling all parameters similarly, we approximate the expectation over the variational distribution by averaging multiple samples. Formally, the first term is computed as:

$$\mathbb{E}_q[\log p(\mathcal{D}|W_d^A, W_u^A, W_d^P, W_u^P)] = \frac{1}{S} \sum_{i=1}^{S} \sum_{(x,y) \in \mathcal{D}} \log p(y|x, W_d^{A(i)}, W_u^{A(i)}, W_d^{P(i)}, W_u^{P(i)}) \tag{25}$$

$S$ is the number of samples in Monte Carlo. $W_d^{A(i)}, W_u^{A(i)}, W_d^{P(i)}, W_u^{P(i)}$ denote the $i$-th sampled parameter matrices via the reparameterization trick. In classification tasks, this log-likelihood typically corresponds to the negative cross-entropy loss; in regression tasks which usually assumes Gaussian noise in the output, this becomes the negative mean squared error (MSE) loss.

### 4.2.3 The Second Term of the Objective

The second term of ELBO is the KL divergence between the variational and the prior distribution. Since the KL divergence for different parameter matrices follows the same computational pattern, we illustrate the derivation by taking $q(W_d^A)$ and $p(W_d^A)$ as an example. For the prior, we adopt the diagonal covariance matrix for the Gaussian distribution. Formally,

$$p(W_d^A) = \mathcal{N}\left(W_d^A; \mu_0, \text{diag}(\sigma_0^2)\right) = \prod_{ij} \mathcal{N}(W_{d,ij}^A; \mu_0, \sigma_0^2) \tag{26}$$

where $\mu_0$ and $\sigma_0$ are prior mean and standard deviation. $\mu_0$ is set to $0$ and each element of $\sigma_0$ is set as 0.1 according to previous studies (Blundell et al. 2015).

We also adopt the diagonal covariance matrix for the variational distribution, i.e., $\Sigma_d^A = \text{diag}\left((\sigma_d^A)^2\right)$. Then, KL divergence between the two Gaussian distributions can be computed as:

$$\text{KL}(q(\boldsymbol{W}_\text{d}^\text{A})\|p(\boldsymbol{W}_\text{d}^\text{A}))$$
$$= \frac{1}{2}\left[\log\frac{|\text{diag}(\boldsymbol{\sigma}_0^2)|}{\left|\text{diag}\left((\boldsymbol{\sigma}_\text{d}^\text{A})^2\right)\right|} - K + (\boldsymbol{\mu}_0 - \boldsymbol{\mu}_\text{d}^\text{A})^\text{T}\text{diag}(\boldsymbol{\sigma}_0^2)^{-1}(\boldsymbol{\mu}_0 - \boldsymbol{\mu}_\text{d}^\text{A}) \right. \tag{27}$$
$$\left. + \text{tr}\left(\text{diag}\left((\boldsymbol{\sigma}_\text{d}^\text{A})^2\right)^{-1}\text{diag}\left((\boldsymbol{\sigma}_\text{d}^\text{A})^2\right)\right)\right]$$

where $K$ is the dimensionality of $\boldsymbol{W}_\text{d}^\text{A}$, (i.e., $K = d \times r_\text{A}$ for a matrix $\boldsymbol{W}_\text{d}^\text{A} \in \mathbb{R}^{d \times r_\text{A}}$). The other KL divergences can be computed similarly.

### 4.2.4 The Optimization Process

After computing both terms of the ELBO, we can determine the value of the ELBO, which is then maximized to learn the variational parameters: $\boldsymbol{\theta} = \{\boldsymbol{\mu}_\text{d}^\text{A}, \boldsymbol{\sigma}_\text{d}^\text{A}, \boldsymbol{\mu}_\text{u}^\text{A}, \boldsymbol{\sigma}_\text{u}^\text{A}, \boldsymbol{\mu}_\text{d}^\text{P}, \boldsymbol{\sigma}_\text{d}^\text{P}, \boldsymbol{\mu}_\text{u}^\text{P}, \boldsymbol{\sigma}_\text{u}^\text{P}\}$. Similar to other PEFT methods, we fix the original model parameters and only update the newly introduced parameters (i.e., the variational parameters in our study). Using standard optimizers (e.g., Adam), we compute the gradient of the negative ELBO with respect to $\boldsymbol{\theta}$ and update $\boldsymbol{\theta}$ via gradient descent.

In order to accelerate convergence, we follow Wang et al. (2025) where the mean $\boldsymbol{\mu}$ is directly parameterized while the standard deviations $\boldsymbol{\sigma}$ is parameterized as $\sigma_{ij} = g_{ij}^2$. We denote $\boldsymbol{g}$ as the collection of $g_{ij}$. Then, $\boldsymbol{\mu}$ and $\boldsymbol{g}$ are initialized as $\boldsymbol{\mu} \sim \text{uniform}\left(-\sqrt{6/d}, \sqrt{6/d}\right)$ and $\boldsymbol{g} \sim \text{uniform}(\delta/\sqrt{2}, \delta)$, where $d$ is the model's hidden dimension (same as that in Section 4.4.1) and $\delta$ is a hyperparameter. The pseudocode of the learning process is shown in Appendix C.

## 5. Our Proposed Bayesian Dynamic Fine-tuning Approach

To address the challenge of newly emerging data in real-world settings, a Bayesian dynamic approach is proposed to enables dynamic fine-tuning based on our BH-PEFT method. Specifically, after obtaining the posterior (i.e., the learned variational distribution) from the current data, we use it as the prior for the next fine-tuning round. With new data, we update the posterior based on this prior. By treating the last posterior as the new prior and then getting the updated posterior, we can apply BH-PEFT continuously to account for new data. The process is shown in Figure 3.

**Figure 3. The Proposed Bayesian Dynamic Fine-tuning Approach**

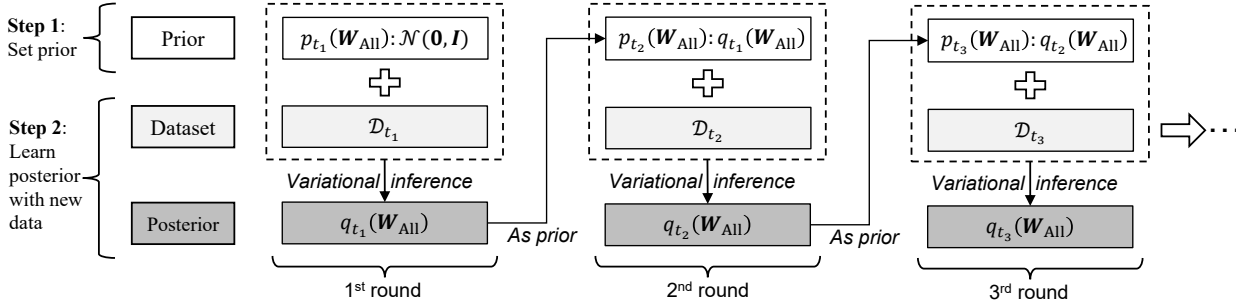

Formally, consider a stream of datasets $\{\mathcal{D}_{t_k}\}_{k=1}^{K}$, where $t_k$ denotes the time period at which the dataset $\mathcal{D}_{t_k}$ is collected. We hope to dynamically process the data stream and infer the distribution $p(\boldsymbol{W}_{\text{All}}|\mathcal{D}_{t_1:t_k})$ at each time $t_k$, i.e., the posterior of all the learnable parameters $\boldsymbol{W}_{\text{All}}$ given the data from time $t_1$ to $t_k$. At time $t_k$, we update $\boldsymbol{W}_{\text{All}}$ using the following two steps.

**Step1 (Set Prior):** At the initial time $t_1$, we set the prior $p_{t_1}(\boldsymbol{W}_{\text{All}})$ as the standard Gaussian. For subsequent time $k > 1$, the prior is set as the posterior obtained at the last time:

$$p_{t_k}(\boldsymbol{W}_{\text{All}}) \leftarrow q_{t_{k-1}}(\boldsymbol{W}_{\text{All}}) \tag{28}$$

$q_{t_{k-1}}(\boldsymbol{W}_{\text{All}})$ is the posterior learned from the variational inference process in the ($k$-1)-th time.

**Step 2 (Update Posterior with New Data):** Given the new dataset $\mathcal{D}_{t_k}$ at time $t_k$, we learn the new posterior $q_{t_k}(\boldsymbol{W}_{\text{All}})$ by repeating the variational inference process mentioned above.

Compared to the existing dynamic fine-tuning approaches as we reviewed in the literature, our method offers three key advantages. First, unlike the data-pooling-based methods which repeat training on the entire pooled dataset, our approach uses the learned posterior as the prior for new data, enabling efficient fine-tuning. Second, unlike parameter-initialization-based methods which simply reuse the previously trained parameters for initialization, our approach uses regularization of prior to constrain the entire fine-tuning process. By requiring the closeness between the posterior and the prior, our method retains previously learned knowledge in a principled manner, effectively mitigating catastrophic forgetting. Third, unlike data-selection-based methods that rely on heuristically selecting subsets of historical data for fine-tuning, our method directly incorporates prior into the posterior update, eliminating the need for manual data selection. These advantages make our approach more efficient and better suited for newly emerging datasets in the real-world.

# 6. Evaluation
## 6.1 Experiment Setup

Same as previous studies (Pfeiffer et al. 2020), we evaluate our BH-PEFT method to fine-tune the RoBERTa-base model. The evaluation covers four business tasks: customer satisfaction prediction, sentiment analysis, news categorization, and commonsense reasoning. Specifically, the datasets include: (1) the Drug Reviews dataset (Gräßer et al. 2018), a text regression dataset where the task is to predict customer satisfaction scores (1–10) based on online drug reviews; (2) the IMDB dataset (Maas et al. 2011), a binary text classification dataset where the task is to determine whether the sentiment of movie reviews is positive or negative; (3) the AG's News dataset (Zhang et al. 2015), a four-class news classification dataset covering the categories *World*, *Sports*, *Business*, and *Technology*; (4) the SST2 dataset (Socher et al. 2013), a sentence-level sentiment classification dataset containing positive and negative online reviews; (5) the CSQA dataset (Talmor et al. 2019), a multiple-choice commonsense reasoning dataset for assessing real-world knowledge and logical inference abilities. A brief summary of these datasets is in Appendix D.

The experiments were conducted using the Hugging Face Transformers library (Wolf et al. 2020), ensuring compatibility with state-of-the-art implementations. The bottleneck dimensions of our method were set at $r_A = 8$ and $r_P = 8$. The scaling factor $s$ was 4. More details are provided in Appendix E. To ensure fair comparisons with other PEFT methods, we adjusted the bottleneck dimensions of the baselines so that the total number of learnable parameters remained comparable.

### 6.2 Experiment Results

#### 6.2.1 Comparison with Single-Aspect-Based PEFT Methods

Experiment 1 evaluated the performance of our BH-PEFT against several existing single-aspect-based PEFT methods, including Adapter (Houlsby et al. 2019), LoRA (Hu et al. 2022) and prefix-tuning (Li and Liang 2021). We also considered several more advanced single-aspect-based PEFT methods such as APT (Zhang et al. 2023), COMPACTER (Karimi Mahabadi et al. 2021), and RandLoRA (Albert et al. 2025). These baselines are widely used, and this comparison aims to demonstrate the advantages of our method over existing single-aspect-based methods.

For regression tasks, we report Mean Squared Error (MSE) and Mean Absolute Error (MAE) to measure model performances. For classification and multiple-choice tasks, we use Accuracy (Acc.), Precision (Pre.), Recall (Rec.), and F1-score (F1) for a comprehensive evaluation. For all datasets, we report the mean and standard deviation across five runs.

Table 1 presents the comparison results of our BH-PEFT method against the baselines. Our method consistently outperforms all baselines across all datasets and evaluation metrics. For instance, on the customer satisfaction prediction task with Drug Reviews dataset, BH-PEFT achieves the lowest MSE and MAE, reducing MSE by 5.68% (i.e., (0.176 – 0.166)/0.176×100%) and MAE by 3.57% compared to the best-performing baseline (COMPACTER). These improvements are statistically significant (p-value < 0.05). Similarly, BH-PEFT also achieves the best performance on sentiment analysis (IMDB, SST2), news categorization (AG's News), and commonsense reasoning (CSQA). These results demonstrate that BH-PEFT outperforms existing single-aspect-based PEFT methods across a range of tasks.

**Table 1. Comparison Between Our BH-PEFT and Single-Aspect-Based PEFT Baselines**

|  | Drug Reviews | | IMDB | | | | AG's News | | | |
|---|---|---|---|---|---|---|---|---|---|---|
| **Methods** | MSE | MAE | Acc. | Pre. | Rec. | F1 | Acc. | Pre. | Rec. | F1 |
| LoRA | 0.203* ±0.005 | 0.297* ±0.001 | 0.952* ±0.002 | 0.952* ±0.002 | 0.952* ±0.002 | 0.952* ±0.002 | 0.939* ±0.001 | 0.939* ±0.001 | 0.938* ±0.001 | 0.938* ±0.002 |
| Prefix-tuning | 0.219** ±0.003 | 0.314** ±0.002 | 0.943* ±0.007 | 0.943* ±0.001 | 0.942* ±0.001 | 0.942* ±0.001 | 0.924** ±0.002 | 0.925** ±0.001 | 0.924** ±0.001 | 0.925** ±0.001 |
| Adapter | 0.179 ±0.009 | 0.272 ±0.003 | 0.954* ±0.002 | 0.955* ±0.001 | 0.954* ±0.001 | 0.954* ±0.001 | 0.943* ±0.005 | 0.945* ±0.002 | 0.944* ±0.001 | 0.945* ±0.002 |
| RandLoRA | 0.176* ±0.008 | 0.281** ±0.004 | 0.956 ±0.002 | 0.957 ±0.002 | 0.956 ±0.002 | 0.956 ±0.002 | 0.950 ±0.001 | 0.951 ±0.001 | 0.950 ±0.001 | 0.950 ±0.001 |
| APT | 0.177** ±0.001 | 0.282* ±0.003 | 0.954 ±0.002 | 0.955 ±0.002 | 0.954 ±0.001 | 0.955 ±0.001 | 0.948 ±0.001 | 0.948 ±0.001 | 0.948 ±0.001 | 0.948 ±0.001 |
| COMPACTER | 0.176* ±0.004 | 0.280* ±0.005 | 0.955 ±0.002 | 0.956 ±0.001 | 0.956 ±0.002 | 0.956 ±0.002 | 0.945* ±0.001 | 0.945* ±0.001 | 0.944* ±0.001 | 0.944* ±0.001 |
| **BH-PEFT** | **0.166** ±0.002 | **0.270** ±0.001 | **0.957** ±0.001 | **0.958** ±0.001 | **0.957** ±0.001 | **0.957** ±0.001 | **0.950** ±0.002 | **0.951** ±0.001 | **0.950** ±0.001 | **0.950** ±0.001 |

|  | SST2 | | | | CSQA | | | |
|---|---|---|---|---|---|---|---|---|
| **Methods** | Acc. | Pre. | Rec. | F1 | Acc. | Pre. | Rec. | F1 |
| LoRA | 0.942* ±0.001 | 0.942* ±0.001 | 0.941* ±0.002 | 0.942* ±0.002 | 0.615* ±0.002 | 0.616* ±0.001 | 0.615* ±0.001 | 0.615* ±0.001 |
| Prefix-tuning | 0.929** ±0.001 | 0.928** ±0.002 | 0.928** ±0.002 | 0.928** ±0.002 | 0.602** ±0.001 | 0.603** ±0.001 | 0.602** ±0.001 | 0.602** ±0.001 |
| Adapter | 0.942 ±0.002 | 0.943 ±0.001 | 0.943 ±0.001 | 0.943 ±0.001 | 0.614* ±0.000 | 0.615* ±0.001 | 0.614* ±0.001 | 0.614* ±0.001 |
| RandLoRA | 0.945 ±0.001 | 0.945 ±0.001 | 0.945 ±0.001 | 0.945 ±0.001 | 0.606* ±0.003 | 0.606* ±0.002 | 0.606* ±0.001 | 0.606* ±0.001 |
| APT | 0.946 ±0.002 | 0.946 ±0.002 | 0.946 ±0.003 | 0.946 ±0.002 | 0.617* ±0.002 | 0.618* ±0.001 | 0.616* ±0.001 | 0.617* ±0.001 |
| COMPACTER | 0.943* ±0.001 | 0.944* ±0.002 | 0.944* ±0.002 | 0.944* ±0.002 | 0.605** ±0.001 | 0.606** ±0.001 | 0.605** ±0.002 | 0.605** ±0.001 |
| **BH-PEFT** | **0.948** ±0.003 | **0.948** ±0.002 | **0.947** ±0.001 | **0.948** ±0.001 | **0.628** ±0.002 | **0.628** ±0.002 | **0.627** ±0.001 | **0.627** ±0.001 |

Notes: * $p < 0.05$, ** $p < 0.01$ (versus BH-PEFT, based on t-test)

### 6.2.2 Comparison with Hybrid-Aspect-Based PEFT Methods

We also compared our method with existing hybrid-aspect-based PEFT methods, including MAM Adapter (He et al. 2022) and UniPELT (Mao et al. 2021). Although these methods also integrate the aspects of multiple distinct PEFT methods such as Adapter, LoRA, and prefix-tuning, they model parameters as deterministic points rather than distributions. This experiment aims to demonstrate the advantage of modeling parameters as distributions through Bayesian learning.

Table 2 presents the comparison results between BH-PEFT and these hybrid PEFT baselines. On the Drug Reviews dataset, BH-PEFT achieves the lowest MSE and MAE, reducing MSE by 4.60% and MAE by 3.23% compared to the best-performing baseline (UniPELT). Both improvements are statistically significant (p-value < 0.01). BH-PEFT likewise delivers the best performance on sentiment analysis (IMDB, SST2), news categorization (AG's News), and commonsense reasoning (CSQA) with statistically significance (p-value < 0.05). This further shows the benefits of distribution-based modeling via Bayesian learning for hybrid PEFT methods.

Table 2. Comparison Between Our BH-PEFT and Hybrid-Aspect-Based PEFT Baselines

| | Drug Reviews | | IMDB | | | | AG's News | | | |
|---|---|---|---|---|---|---|---|---|---|---|
| Methods | MSE | MAE | Acc. | Pre. | Rec. | F1 | Acc. | Pre. | Rec. | F1 |
| UniPELT | 0.174** ±0.004 | 0.279** ±0.004 | 0.955 ±0.002 | 0.955 ±0.001 | 0.955 ±0.002 | 0.955 ±0.002 | 0.943* ±0.001 | 0.942* ±0.001 | 0.943* ±0.001 | 0.943* ±0.001 |
| MAM Adapter | 0.177** ±0.002 | 0.277** ±0.001 | 0.954** ±0.001 | 0.954** ±0.002 | 0.954** ±0.002 | 0.954** ±0.002 | 0.943** ±0.002 | 0.945** ±0.002 | 0.944** ±0.002 | 0.944** ±0.002 |
| BH-PEFT | **0.166** ±0.002 | **0.270** ±0.001 | **0.957** ±0.001 | **0.958** ±0.001 | **0.957** ±0.001 | **0.957** ±0.001 | **0.950** ±0.002 | **0.951** ±0.001 | **0.950** ±0.001 | **0.950** ±0.001 |

| | SST2 | | | | CSQA | | | |
|---|---|---|---|---|---|---|---|---|
| Methods | Acc. | Pre. | Rec. | F1 | Acc. | Pre. | Rec. | F1 |
| UniPELT | 0.943* ±0.002 | 0.943* ±0.001 | 0.943* ±0.001 | 0.943* ±0.001 | 0.612** ±0.002 | 0.613** ±0.002 | 0.612** ±0.002 | 0.612** ±0.002 |
| MAM Adapter | 0.943** ±0.001 | 0.943* ±0.001 | 0.943* ±0.001 | 0.943* ±0.001 | 0.620** ±0.001 | 0.621** ±0.001 | 0.620** ±0.000 | 0.620** ±0.000 |
| BH-PEFT | **0.948** ±0.003 | **0.948** ±0.002 | **0.947** ±0.001 | **0.948** ±0.001 | **0.628** ±0.002 | **0.628** ±0.002 | **0.627** ±0.001 | **0.627** ±0.001 |

### 6.2.3 Ablation Analysis

We conducted an ablation study to assess the contribution of the key designs of this study. By removing each design and comparing the model's performance with and without it, we identify its impact, where a performance drop indicates its effectiveness. Specifically, we evaluated three variants: 1) replacing Bayesian learning with conventional deterministic point-based modeling; 2) disabling Bayesian prefix-tuning in the multi-head self-attention layer; and 3) removing the Bayesian scaled parallel adapter (combining LoRA and Adapter) from the feedforward layers.

As shown in Table 3, all variants attain inferior results, highlighting the value of each design. First, when Bayesian learning is replaced with point-based fine-tuning, the model consistently underperforms across all tasks. For instance, on the Drug Reviews dataset, MSE increases by 6.63% and MAE increases by 3.86%, and shows a statistically significant degradation (p-value < 0.01). This indicates the distribution-based modeling is more effective than the deterministic point-based approach. Second, disabling Bayesian prefix-tuning leads to significant drops. For instance, on the CSQA dataset, accuracy decreases by 3.03% with statistically significance (p-value < 0.01). Other metrics similarly deteriorate, indicating that the Bayesian prefix-tuning plays an essential role in enhancing model's performance. Third, removing the Bayesian scaled parallel adapter results in performance drops. For instance, on the AG's News dataset, this removal results in a significant drop in all metrics (p-value < 0.01), highlighting the importance of Bayesian scaled parallel adapter within the feedforward layers. Together, these findings confirm that both Bayesian prefix-tuning and Bayesian scaled parallel adapter make complementary and significant contributions, and that their integration enables BH-PEFT to achieve consistent gains across diverse specialized task.

Table 3. Results of the Ablation Analysis

| Methods | Drug Reviews | | IMDB | | | | AG's News | | | |
|---|---|---|---|---|---|---|---|---|---|---|
| | MSE | MAE | Acc. | Pre. | Rec. | F1 | Acc. | Pre. | Rec. | F1 |
| No Bayesian Learning | 0.177** ±0.002 | 0.278** ±0.001 | 0.954** ±0.001 | 0.955** ±0.002 | 0.954** ±0.003 | 0.955** ±0.003 | 0.943** ±0.002 | 0.944** ±0.002 | 0.943** ±0.002 | 0.944** ±0.002 |
| Without Bayesian Scaled Parallel Adapter | 0.176** ±0.002 | 0.282** ±0.003 | 0.954** ±0.002 | 0.954** ±0.002 | 0.954** ±0.002 | 0.954** ±0.002 | 0.944** ±0.002 | 0.943** ±0.001 | 0.942** ±0.001 | 0.943** ±0.001 |
| Without Bayesian Prefix-Tuning | 0.174** ±0.001 | 0.278** ±0.003 | 0.956 ±0.001 | 0.955 ±0.002 | 0.955 ±0.002 | 0.955 ±0.002 | 0.947* ±0.002 | 0.947* ±0.001 | 0.947* ±0.002 | 0.947* ±0.002 |
| Full BH-PEFT | **0.166** ±0.002 | **0.270** ±0.001 | **0.957** ±0.001 | **0.958** ±0.001 | **0.957** ±0.001 | **0.957** ±0.001 | **0.950** ±0.002 | **0.951** ±0.001 | **0.950** ±0.001 | **0.950** ±0.001 |

| Methods | SST2 | | | | CSQA | | | |
|---|---|---|---|---|---|---|---|---|
| | Acc. | Pre. | Rec. | F1 | Acc. | Pre. | Rec. | F1 |
| No Bayesian Learning | 0.943** ±0.001 | 0.943** ±0.002 | 0.942** ±0.001 | 0.942** ±0.001 | 0.620** ±0.001 | 0.620** ±0.001 | 0.620** ±0.000 | 0.620** ±0.000 |
| Without Bayesian Scaled Parallel Adapter | 0.943 ±0.005 | 0.945 ±0.003 | 0.945 ±0.003 | 0.945 ±0.003 | 0.616** ±0.001 | 0.617** ±0.002 | 0.616** ±0.002 | 0.616** ±0.002 |
| Without Bayesian Prefix-Tuning | 0.944* ±0.002 | 0.944* ±0.001 | 0.944* ±0.001 | 0.944* ±0.001 | 0.609** ±0.001 | 0.609** ±0.001 | 0.608** ±0.001 | 0.608** ±0.001 |
| Full BH-PEFT | **0.948** ±0.003 | **0.948** ±0.002 | **0.947** ±0.001 | **0.948** ±0.001 | **0.628** ±0.002 | **0.628** ±0.002 | **0.627** ±0.001 | **0.627** ±0.001 |

### 6.2.4 Analyzing the Quantified Uncertainty Using Rejection Curve

A key advantage of our method is its capacity for uncertainty quantification. Hence, we next demonstrate the benefits of the quantified uncertainty. We adopted rejection curve (Nadeem et al. 2009), which reflects how uncertainty correlates with the error in the model outputs. Specifically, a rejection curve plots the model's performance (e.g., accuracy or MSE) as a function of the rejection rate, where predictions with the highest uncertainty are progressively filtered out (i.e., rejected). Since high uncertainty implies the prediction lacks confidence and is unreliable, we are expected to attain more accurate predictions in low-uncertainty examples.

Figure 3 presents the rejection curves, illustrating how performance improves as predictions with the highest uncertainty are progressively rejected across the specialized tasks. For instance, in the customer satisfaction prediction task with the Drug Reviews dataset, the MSE consistently decreases as the rejection rate increases. Notably, excluding just the top 10% most uncertain predictions reduces MSE by 18.45%. In the news categorization task on the AG News dataset, rejecting the top 10% most uncertain samples improves accuracy by approximately 3%, and this upward trend continues with higher rejection thresholds. This pattern shows that our uncertainty quantification helps flag unreliable outputs and thus can bring about more reliable decision-making.

**Figure 3. Rejection Curves Across Different Datasets**

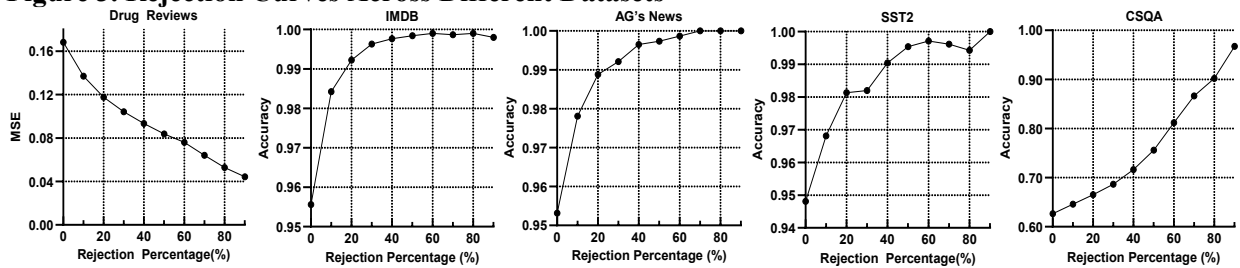

**6.2.5 Experiment of Dynamic Fine-Tuning**

This experiment aims to showcase the advantages of our Bayesian approach in dynamic fine-tuning. Due to the relatively high computational cost, the experiment was conducted on a single dataset (SST-2). The training began with 10 samples and progressively increased the dataset size in a geometric progression (i.e., 10, 20, 40, ..., up to 5120) over 10 rounds, simulating realistic scenarios of incremental data availability. We compared our method with three other types of baselines including data-pooling-based (Garg et al. 2024), parameter-initialization-based (Soutif et al. 2024), and data-selection-based methods (Song et al. 2023).

As shown in Table 4, our Bayesian approach performs on par with the baselines in early rounds, but shows a clear upward trend from the fifth round onward. In the final round (i.e., the 10th round), our approach achieves an accuracy of 0.942, significantly outperforming the other three baselines (p-value < 0.01). On average, it achieves an accuracy of 0.867, showing a relative improvement of 0.6% to 3.1% over the baselines. These results suggest that our BH-PEFT offers a more effective approach for dynamic fine-tuning to cope with the newly emerging dataset.

**Table 4. Experiment Results of Dynamic Fine-Tuning on SST-2 Dataset**

| Round | Our Bayesian Approach | Data-Pooling-Based | Parameter-Initialization-Based | Data-Selection-Based |
|---|---|---|---|---|
| 1 | **0.653**±0.003 | 0.652±0.004** | 0.650±0.001** | 0.641±0.009** |
| 2 | 0.782±0.001 | **0.783**±0.002 | 0.737±0.001 | 0.778±0.003 |
| 3 | 0.824±0.001 | 0.827±0.007 | 0.814±0.000 | **0.842**±0.002 |
| 4 | **0.865**±0.001 | 0.854±0.013 | 0.811±0.004** | 0.858±0.005* |
| 5 | **0.907**±0.001 | 0.881±0.001** | 0.836±0.002** | 0.884±0.001** |
| 6 | **0.918**±0.001 | 0.907±0.002** | 0.884±0.001** | 0.910±0.002** |
| 7 | **0.922**±0.003 | 0.922±0.001 | 0.903±0.002** | 0.921±0.001 |
| 8 | **0.926**±0.002 | 0.922±0.000* | 0.913±0.001** | 0.923±0.001* |
| 9 | **0.934**±0.002 | 0.930±0.002* | 0.925±0.002** | 0.929±0.002** |
| 10 | **0.942**±0.001 | 0.933±0.002** | 0.933±0.004** | 0.934±0.002** |
| Avg | **0.867** | 0.861* | 0.841** | 0.862 |

## 7. Discussion and Conclusion

LLMs have emerged as a transformative force for various business applications, and PEFT methods have been instrumental in enhancing the capabilities of LLMs. While various methods have been proposed, such as the Adapter-based methods, LoRA-based methods, and prefix-tuning-based methods, hybrid aspect-based methods have achieved better performance. However, the existing hybrid PEFT methods are still limited in that they rely on point estimates and thus fail to reflect uncertainty, and they struggle to dynamically adapt to new data. Hence, we proposed Bayesian Hybrid Parameter-Efficient Fine-Tuning (BH-PEFT), a novel fine-tuning method that introduces Bayesian learning into hybrid PEFT methods. By treating the learnable parameters as random variables from probabilistic distributions, BH-PEFT enhances hybrid methods by allowing for uncertainty quantification and better adaptability to new data. Extensive experiments demonstrate the effectiveness of our method on various business tasks including customer satisfaction prediction, sentiment analysis, news categorization, and commonsense reasoning.

### 7.1 Research Implications

This study has important implications for advancing PEFT research on LLMs. While existing hybrid PEFT methods combine multiple fine-tuning aspects, they typically rely on deterministic point-based modeling, limiting their ability to capture uncertainty or retain previously learned knowledge during dynamic updates. Through Bayesian learning, our method models learnable parameters as probability distributions, enabling uncertainty quantification via the variance of the posterior distribution. This enables researchers to assess the model confidence and flag the low-reliability predictions, leading to more reliable LLM-assisted decision-making. In addition, the iterative use of posterior distributions as the priors for future updates aligns well with the need for dynamic fine-tuning, which is crucial for adapting to new data while avoiding catastrophic forgetting. These designs not only enhance the reliability and effectiveness of PEFT methods but also contribute to the broader understanding of how Bayesian learning can be integrated into modular and hybrid fine-tuning approaches. Our study thus establishes a foundation for further exploring the intersection between Bayesian learning and PEFT, offering valuable insights for developing more reliable and adaptable LLMs for specialized applications.

### 7.2 Managerial Implications

This study offers important managerial implications for developing LLMs in real-world business applications. We focus on two key perspectives: uncertainty quantification and dynamic fine-tuning. First, uncertainty quantification enhances the reliability of AI-assisted decision-making. In many real-world situations, managers use LLM outputs to guide operational or strategic decisions. However, conventional LLMs provide only point-based predictions without indicating how confident the model is, which may lead to overreliance on uncertain results and increased decision risk. For example, Yu (2023) shows that sentiment-based forecasting models without uncertainty control can lead to unstable investment strategies due to unreliable predictions, while incorporating uncertainty helps identify low-confidence outputs and improves decision reliability. BH-PEFT tackles this problem by integrating uncertainty quantification into fine-tuning, enabling practitioners to measure the confidence and reliability of model outputs. This directly impacts how organizations manage AI-assisted workflows, allowing managers to apply LLMs more cautiously and transparently by setting confidence thresholds for human review or filtering out unreliable outputs. Second, our Bayesian dynamic fine-tuning helps LLMs adapt to new data. In practice, new data is continuously generated through user behavior, market dynamics, or internal operations. However, existing fine-tuning methods are limited by high computational costs, catastrophically forgetting previously learned knowledge, or relying on subjective data selection. BH-PEFT provides a better solution through Bayesian learning, allowing the model to integrate new information while retaining prior knowledge. This avoids full retraining, reduces the risk of catastrophic forgetting, and eliminates subjective biases that could harm performance. Our approach is especially valuable in business areas such as financial services, cybersecurity, operational management, and community question-answering, where domain knowledge is rich and crucial. In summary, our BH-PEFT method and the dynamic fine-tuning approach based on it help organizations improve the reliability of AI-assisted decision-making while effectively adapting their LLMs to new data, keeping them up-to-date.

### 7.3 Limitations and Future Research

As with other studies, our study also has some limitations and corresponding directions. First, due to limited computational resources, this work primarily focuses on supervised fine-tuning settings such as classification and regression tasks. While BH-PEFT may also apply to more complex tasks like multi-task learning and open-ended generation, uncertainty quantification in these areas differs and requires further investigation. Future research can explore the effectiveness of our approach in such settings. Second, due to computational constraints, we adopt a relatively lightweight model (i.e., RoBERTa-base) in our experiments. Future work can explore the performance of our method when applied to larger-scale models. Third, our current dynamic fine-tuning approach primarily considers the scenario of a single stream of continuously arriving data. A possible future direction is to extend the method to multi-task or cross-domain dynamic fine-tuning scenarios, where multiple data streams may pose challenges to dynamic fine-tuning.